\documentclass{article}

\usepackage[square,sort,comma,numbers]{natbib}

\usepackage[preprint]{neurips_2023}
\usepackage{epsfig}
\usepackage{graphicx}
\usepackage{amsmath}
\usepackage{amssymb}
\usepackage{algorithm} 
\usepackage{algpseudocode}
\usepackage{multirow}
	
\graphicspath{ {./pics/} }

\usepackage[utf8]{inputenc} 
\usepackage[T1]{fontenc}    
\usepackage{hyperref}       
\usepackage{url}            
\usepackage{booktabs}       
\usepackage{amsfonts}       
\usepackage{nicefrac}       
\usepackage{microtype}      
\usepackage{xcolor}         

\usepackage{diagbox}

\title{Self-supervised novel 2D view synthesis of large-scale scenes with efficient multi-scale voxel carving}

\author{%
  Alexandra Budi\c{s}teanu\thanks{Alexandra Budi\c{s}teanu is with the Institute of Mathematics of the Romanian Academy.} \\
  \texttt{budisteanu.alexandra@gmail.com} \\
  \And
  Drago\c{s} Costea\thanks{Drago\c{s} Costea is with the Institute of Mathematics of the Romanian Academy and University Politehnica of Bucharest.}\\
  \texttt{dragos.costea@upb.ro} \\
  \And
  Alina Marcu\thanks{Alina Marcu is with the Institute of Mathematics of the Romanian Academy and University Politehnica of Bucharest.}\\
  \texttt{alina.marcu@upb.ro} \\
  \And
  Marius Leordeanu\thanks{Marius Leordeanu is the corresponding author and is with the Institute of Mathematics of the Romanian Academy and University Politehnica of Bucharest.} \\
  \texttt{marius.leordeanu@upb.ro} \\
}

\begin{document}

\maketitle

\begin{abstract}
The task of generating novel views of real scenes is increasingly important nowadays when AI models become able to create realistic new worlds. In many practical applications, it is important for novel view synthesis methods to stay grounded in the physical world as much as possible, while also being able to imagine it from previously unseen views. While most current methods are developed and tested in virtual environments with small scenes and no errors in pose and depth information, we push the boundaries to the real-world domain of large scales in the new context of UAVs. Our algorithmic contributions are two folds. First, we manage to stay anchored in the real 3D world, by introducing an efficient multi-scale voxel carving method, which is able to accommodate significant noises in pose, depth, and illumination variations, while being able to reconstruct the view of the world from drastically different poses at test time. Second, our final high-resolution output is efficiently self-trained on data automatically generated by the voxel carving module, which gives it the flexibility to adapt efficiently to any scene. We demonstrated the effectiveness of our method on highly complex and large-scale scenes in real environments while outperforming the current state-of-the-art. Our code is publicly available: \url{https://github.com/onorabil/MSVC}.
\end{abstract}

\section{Introduction}
\label{sec:intro}

Current AI models are becoming increasingly capable to generate realistic fake worlds, which may be harmful in applications where we need to stay grounded in the real physical world. At the same time, in many tasks, it is also essential to be able to imagine how the world really looks from different completely novel views, with minimal use of resources and computation cost. For example, in many cases (e.g. creating artistic movies and other kinds of visual content) the ability to imagine virtual flights over a real scene could be of great practical value, but without modifying the true structure of the real scene.

Prior work for novel view synthesis focused only on small synthetic scenes, with zero noise in input data (for pose or depth) and only with very small variations in novel poses versus the ones seen in training. This is a very limited scenario, with very little use in the real world and here, we come to address this limitation. Our goal is to address all these limitations, along several dimensions, which also define our main contributions.

We address the problem of grounding in the 3D structure of the real world, with a novel and very efficient multi-scale voxel carving method, which considers voxels at different scales (sizes) and decides upon their 3D existence (part of the real surface in the world) and color using a robust confidence voting-based measure. Our multi-scale voxelization is not used to explicitly build the 3D structure of the world, but only to minimize 2D view reconstruction error and coverage from novel unseen camera poses. This is  different from current 2D synthesis methods, which are not based on a real 3D structure, and also different from explicit 3D reconstruction methods, which are accurate but have very weak 2D reconstruction coverage, being limited only to poses seen during training.

We address the second important aspect of generating accurate and high quality (resolution) 2D views of a real scene, by introducing a second neural net module, which learns automatically supervised by the first module, on the training views of the same scene, such that it can automatically adapt to any scene, for maximum efficiency. This module essentially consists of a small U-Net, which is rapidly fine-tuned during the multi-scale voxel carving stage and learns to reconstruct the original RGB from the output of the Multi-Scale Voxel-Carving (MSVC) based 2D view reconstruction, at minimal additional cost.

Thus, the MSVC teacher module grounds the reconstruction in the real world, being robust to noises coming from depth and pose due to its multi-scale voxelization structure, while the second model learns to refine its output and improve coverage during the automatic self-supervised stage.

Our main contributions are summarized below:

\begin{enumerate}
\item We introduce a novel self-supervised system of novel 2D view reconstruction, in which an analytical module, based on multi-scale voxelization, grounds the output into the real 3D world and supervises automatically a second neural net module for accurate high-resolution reconstruction of novel views of the scene.
\item We test and compare our work with state-of-the-art methods, on very difficult cases of real-world large-scale scenes, captured by UAVs, having a significant amount of noise in pose and depth, unlike all other works published so far which are limited to small-scale scenes, synthetic, having no measurement errors and only small variations in view between the train and test cases.
\end{enumerate}




\begin{figure*}
\begin{center}
\includegraphics[scale=0.115]{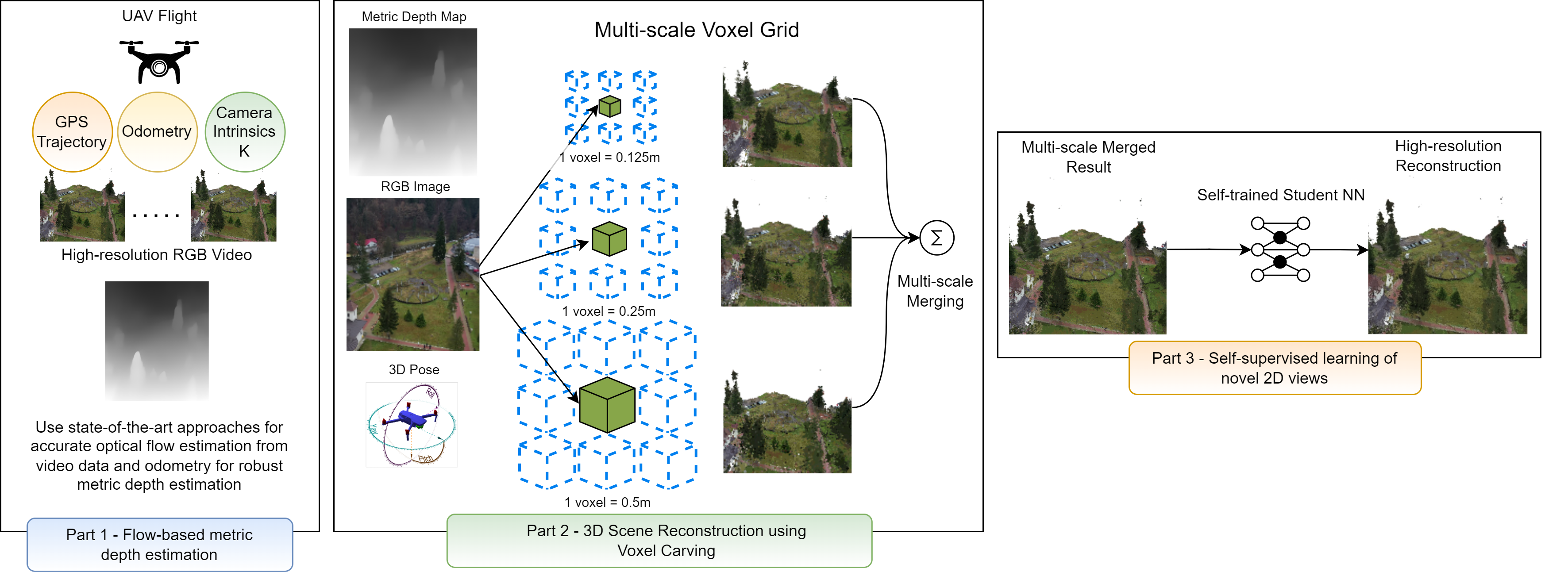}
\caption {We present an overview of our novel self-supervised system of novel 2D views reconstruction. We leverage video and telemetry data from UAV flights and state-of-the-art optical flow-based approaches to obtain accurate metric depth estimation for a given real-world scene. Using RGB, Depth and Pose see apply our novel multi-scale voxelization procedure for 3D reconstruction, which becomes the supervisory signal for a self-supervised neural network in order to obtain accurate high-resolution reconstruction of novel 2D views of the scene.}\label{fig:overview}
\end{center}
\vspace{-4mm}
\end{figure*}

\vspace{-3mm}

\section{Related work}\label{rw}

\textbf{Scene reconstruction} The traditional, feature-based 3D reconstruction structure-from-motion pipelines such as COLMAP~\cite{schoenberger2016sfm} need significant processing time, even for a small number of high-resolution images~\cite{jiang2020efficient}. As the number of images increases, the resources required increase at an exponential rate\cite{triggs2000bundle}, making it unfeasible for large datasets. 

The implicit representation proposed by Neural Radiance Fields (NeRF)~\cite{mildenhall2021nerf} promised an alternative to the compute-intensive structure-from-motion pipelines~\cite{griwodz2021alicevision}. The advent of fast NeRF training frameworks such as Instant Neural Graphics Primitives~\cite{muller2022instant} and NerfAcc~\cite{li2022nerfacc} has reduced the training time from a couple of days for a single scene to a couple of hours or even several minutes. Transforming the input into rays is a shared starting point with our method, but ours has an explicit representation instead. Direct Voxel Grid Optimization~\cite{sun2022direct} proposes a representation that replaces the MLP usually used in NeRFs with a voxel grid, resulting in speed and performance gains. This approach shares both the input and output representation with our method. Nevertheless, most of the time, the reconstructed depth is not dense - marching cubes and a density threshold are needed for actual depth reconstruction. The depth quality and resolved detail are not adapted for large datasets. Generally, the whole dataset is loaded into memory and empirically, the PSNR score from a higher resolution is similar (see our experiments in Section~\ref{sec:comparisons}). 

Furthermore, another downside of these methods is the requirement for accurate pose - structure-from-motion must be run prior to the reconstruction in order to estimate camera poses. This is also a very expensive operation, mostly due to the bundle adjustment optimization that requires significant resources, reaching hundreds of millions of matches. A key contribution of this work is that our method achieves competitive results using only video and telemetry data (from consumer GPS and onboard sensors - no advanced positioning hardware needed). A bundle adjustment NeRF was also developed to address this issue~\cite{chen2022local}.

\textbf{Large scale reconstruction} Given the errors that occur over a large set of images, some authors have proposed various workarounds, usually involving splitting the scene into a set of smaller scenes or grid~\cite{tancik2022block,turki2023suds,xu2023gridguided}, splitting the path into image sets~\cite{meuleman2023localrf}, or splitting the scene into smaller NeRFs~\cite{zhang2023nerflets}. We argue that these methods are not globally consistent and aim to hide the pose and reconstruction errors.


\textbf{Voxel carving} On the other hand, some of the traditional Novel View Synthesis methods relied on a prior scene reconstruction, such as \cite{seitz1999photorealistic}. The authors propose a reconstruction scheme based on a voxelization of the scene and multi-view consistency of the voxels. In their work, the scene is defined as a set of voxels, arranged in such a way that it can be represented such as consecutive layers of voxels. These layers are considered based on their distance from themselves to the camera, meaning that the missing depth information is circumvented by the arrangement of the voxels. Moreover, the method assumes multiple views from a set of cameras. However, the cameras are pre-set in specific positions, resembling the surface of a sphere, with the reconstructed scene in the center of the said sphere. Using the preset positions of the cameras and the layering of the voxels, the proposed algorithm projects voxels into the image space and compares the colors of the voxels with the colors in the input image. If the voxel's color is not consistent across multiple images, the voxel is then carved. A similar approach with fixed cameras is descibed in \cite{balan2003voxel}. Even though modern approaches focus heavily on implicit representations using neural networks, our method aims to improve the classical representations of scene reconstruction using voxelization and novel view synthesis. We introduce the metrics of \textit{depth consistency} and \textit{color consistency}, following mathematical 3D representations of the scene, and also define various optimizations in order to solve some of the issues state-of-the-art methods are facing, improving both the resolution and speed for large-scale datasets.


\section{Self-supervised novel 2D view reconstruction} \label{sec:method}

Inspired by the Voxel Carving~\cite{kutulakos2000theory} method, we develop our algorithm starting from a scene represented by a volume of voxels. We then iteratively carve voxels based on the drone data until all the remaining voxels are consistent across all the viewpoints. We furthermore define two key concepts on which we construct our approach: \textit{depth consistency} and \textit{color consistency}. These two concepts represent properties of the voxels which comprise the final 3D scene.
The data used throughout our experiments consists of a set of \textit{poses} of the drone, a set of \textit{RGB frames}, and their respective \textit{depth  maps}. We refer to the \(pose\) as the position of the drone, expressed in 3D World Coordinates (x, y, z), and its orientation, described as Euler angles. Therefore, we represent our dataset as a set of \(T\) quadruples \(P_{xyz}, E_{xyz}, RGB, Depth\).

\subsection{Backprojection}\label{subsec:backproj}

We define a $voxel$ as the smallest unit in the 3D scene and we represent it as a cube. One cube is defined with a 3D point in the world and the length of its side. We arrange multiple voxels in a grid, such that voxels are directly adjacent to each other, with no gaps in between them. We thus define \(V\) as the set of all initial voxels of the scene \(V = \{v_{xyz} = [v_x, v_y, v_z]\}\) and we refer to it as \textit{Voxel Grid (VG)}. The size of the voxel is considered constant and it is shared across all voxels. The voxel grid is constructed in a way such that by placing a camera in a pose from the dataset, the captured image contains parts of the voxel grid, for every pose in the dataset.

The proposed algorithm aims to carve the voxel grid in such a way that after the carving is complete, the existing voxels satisfy both the depth consistency and color consistency criteria (detailed in Sections~\ref{subsec:depth_cons} and ~\ref{subsec:color_cons}). To this end, we employ an iterative process in which we pass through all the quadruples of items in the dataset and check all voxels from the voxel grid if they describe the aforementioned properties using a \textit{backprojection} from the 3D world space to the 2D image space of the drone's camera.

The backprojection of a voxel from the world space to the image space is done using the position of the voxel in the world space \(v_{xyz}\), the position of the drone \(P_{xyz}\), the orientation of the drone \(E_{xyz}\) and the intrinsic camera parameters of the drone, which we refer to as \(K\). The image space is a two-dimensional space, meaning that the backprojection of a voxel can be represented as a two-dimensional vector, which we call \(v_{proj} = [vp_x, vp_y]\).

Computing the \(v_{proj}\) of a voxel into the image space is done by a double projection. First, we project the voxel from the world space to the camera space, then we project from the camera space to the image space. These two projections can be done in a single step using Equation~\ref{eq:1},

\begin{equation} \label{eq:1}
v_{proj} = K * [(P_{xyz} - v_{xyz}) * R] = [v_{x'}, v_{y'}, v_{z'}]'
\end{equation}

where \(K\) is the intrinsic matrix, \(P_{xyz}\) is the position of the drone in world coordinates, \(v_{xyz}\) is the position of the voxel in world coordinates, and \(R\) is the rotation matrix obtained from the orientation of the drone \(E_{xyz}\). However, the resulting vector requires a further scaling operation to convert to homogeneous coordinates. The scaling factor \(v_{z'}\) also acts as the numerical distance of the voxel from the camera. Thus, we compute \(v_{proj} = [\frac{v_{x'}}{v_{z'}}, \frac{v_{y'}}{v_{z'}}]'\). 

\subsection{Depth consistency} \label{subsec:depth_cons}

The depth consistency property is determined using a voting mechanism employed throughout the reconstruction process. Let \(N = card(V)\) be the number of the voxels from the voxel grid. For each voxel \(v \in V\) we define a \(seen\) measure, which counts how many times a voxel has been \(seen\) from all the viewpoints of the dataset. A voxel is considered to be \(seen\), from a camera at position \(P_i\) having the orientation \(E_i\), if by projecting the voxel from the world space into the image space, its position is within the bounds of the image \(RGB_i\). Moreover, as the depth map \(D_i\) is also available, we strengthen this condition by forcing the voxel to also be approximately on par with the depth information, as we can compute its real distance from the camera and the pixel coordinates in which the voxel projects. 

For the depth consistency part, we extract information from the depth map based on the pixel coordinates of a voxel. In Section~\ref{subsec:backproj} we described how to compute the pixel coordinates of the projection of a voxel in the image space. These pixel coordinates are used to extract information from both the RGB image and its respective depth map. Assuming the voxel projects inside the boundaries of the image, the depth map offers us the expected depth of the voxel, which should be the expected distance from the camera. To this end, we employ Equation \ref{eq:2}, where \(W\) and \(H\) are the width and height, respectively, of the image:

\begin{equation} \label{eq:2}
s_{in}(v)= \left
\{
\begin{array}{ll}
      1,if\ 0 \leq vp_{x} \leq W\ \& \ 0 \leq vp_{y} \leq H\\
      0, otherwise
\end{array} 
\right. 
\end{equation}

Computing the distance from a voxel to the camera has been done during the backprojection step. In the non-homogeneous form of the projection of the voxel, we have the \(v_{z'}\) component which essentially represents the distance between the voxel and the camera. The \(v_{proj}\) vector defines in which pixel the voxel projects, such that using the depth map we extract the expected distance of the voxel from the camera. Thus we denote $ e_{dist}(v) = Depth[v_{proj}] $ and $ c_{dist}(v) = v_{z'}$ which describe the expected distance of a voxel from the camera and the real distance of a voxel from the camera, respectively. We denote another consistency measure for a voxel based on the depth information in Equation \ref{eq:5}, and a parameter \(\varepsilon_{seen}\) which controls the threshold for the depth consistency.

\begin{equation} \label{eq:5}
s_{depth}(v)= \left\{
\begin{array}{ll}
      1,\ if\ |e_{dist}(v) - c_{dist}(v)| < \varepsilon_{seen}\\
      0, otherwise
\end{array} 
\right.
\end{equation}

We define our \textit{depth consistency} function as $seen(v) = s_{in}(v) * s_{depth}(v)$.

The \textit{seen} function describes whether a voxel is depth consistent from a given position and orientation of the camera, based on the corresponding depth map. However, as there are multiple camera poses across the reconstruction data, we define a voting scheme by accumulating the times a voxel is considered as \textit{seen}. To this end, during initialization, we define an array that accumulates the values of the \(seen\) function for all voxels, across all the reconstruction points. We call this array \(seen\_array\) and formally define it as 
$seen\_array(v) = \sum_{i=1}^{T}seen(v)$, where each \(i\) represents a pose of the drone. The voting accumulation allows us to further define a threshold for which a voxel is considered depth consistent across all frames if \(seen\_array(v) > seen\_threshold\), where \(seen\_threshold\) is a parameter. 


\subsection{Color consistency} \label{subsec:color_cons}

If the voxel projects inside the image bounds, we can also get the expected color of the voxel, based on the pixel in which it projects, as we can extract it from the RGB image itself. A voxel is considered color consistent if it can be seen from multiple viewpoints with roughly the same color.

The main objective is to assign colors to each voxel such that they are consistent with the RGB images captured by the drone. In Section~\ref{subsec:depth_cons} we have defined a \(seen\) measure to represent whether a voxel is actually consistent from a given point of view. With this in mind and knowing which pixel, a voxel projects, we can also extrapolate its expected color, by querying the RGB image in that specific position. However, the expected voxel color is bound to the current point of view of the drone, thus we need a way to aggregate all the possible colors from all the viewpoints of the drone.  

Whenever a voxel is projected to the image space, the \textit{seen} function says whether it is actually seen by the camera or not. Furthermore, the exact pixel in which it projects is computed, such that the RGB image is queried for the expected color of the voxel. To counter different illumination variations in the RGB color space, we instead use the HSV color space alongside a discretization scheme. The discretization of the HSV space is done on each channel individually, by splitting the whole interval on which the channel is defined into multiple equal smaller intervals, for which we refer to as \textit{bins}. Whenever we convert a bin back to its HSV counterpart, we select the middle point of the interval as the representative point.

The HSV color space has been discretized into 15 bins on the Hue channel, 10 bins on the Saturation channel, and 10 bins on the Value channel. In total, for each voxel, there can be 1500 combinations of HSV bins for all the colors in the HSV spectrum. Essentially, the discretization method maps any given color of the HSV color spectrum into one of the 1500 bins that we have defined, facilitating 'closer' colors to be mapped in the same bin. This means that a voxel, seen from multiple viewpoints, has slightly different colors due to changes in illumination or small shadows and its color will be mapped in the same bin, further enhancing that the voxel is color consistent. 

Formally, let \(v\) be a voxel for which \(seen(v) = 1\), given a specific position and orientation of the drone. The image \(RGB_i\) is the image captured by the drone in this scenario. As the voxel projects inside the image bounds, \(vp_x\) and \(vp_y\) represent the exact pixel where the voxel projects, leading to \(RGB_i[vp_y,vp_x]\) being the expected RGB color of the voxel. Converting the RGB image to HSV and applying the discretization method, we arrive at Equation \ref{eq:8}.

\begin{equation} \label{eq:8}
bin(v) = discretize(HSV_i[vp_y, vp_x])=[v_H, v_S, v_V]'
\end{equation}

\(v_H\), \(v_S\), and \(v_V\) represent the bin on the Hue, Saturation, and Value components for the original RGB color. In a similar way to how depth consistency is defined, we need a way to aggregate the colors with which the voxels are seen across all the data points. We define an array called \(bins\) which has \(card(V)\) lines and 1500 columns. The number of columns is computed from the number of combinations of the discretized HSV space, for which we use a basic linearization method for the bins themselves. This array is updated through each frame during the reconstruction algorithm for each voxel that is seen, according to the color resembled in the RGB image. 

For depth consistency, we added a single vote for the aggregation to decide whether a voxel is consistent across all the frames. In the case of color consistency, we instead propose a different method, which takes into account both the distance between the voxel and the camera and the difference between the expected depth of the voxel and its actual one. Our intuition comes from the fact that voxels far away from the camera tend to have lower color components in the RGB image, such that we prefer to give more importance to voxels closer to the camera as they appear much bigger. On the other hand, if the expected depth of the voxel is further away than the actual depth of the voxel, it means that the voxel is occluded in the scene, meaning that the current color should not have too much importance on the real color of the voxel.

According to our intuition, we define $f_1(x) = e^{-\frac{ln(\alpha)}{250} * x}$ as a function of distance which is monotonically decreasing with the distance between the voxel and the camera. Voxels farther away will have small values when plugged into this function, while voxels closer to the camera describe higher values. We also define $f_2(x) = e^{\frac{-x^2}{2 * \sigma ^ 2}}$ also monotonically decreasing based on the difference between the expected depth and the computed depth of the voxel. For voxels with similar computed and expected depths, the function has high values.


The \(\alpha\) parameter controls what value should the function have whenever the voxels are close to the maximum distance (of 250 meters in our experiments). We ignore voxels for which their distance is greater than 250 meters as they tend to induce errors during the reconstruction. The \(\sigma\) parameter controls the width around the Gaussian representation, which translates into how large the difference between the expected depth and the computed depth can be. 

Given a voxel \(v\) and a specific pose of the drone, we define the color consistency as in Equation \ref{eq:11}, 

\begin{equation}\label{eq:11}
color(v) = f_1(v_{z'}) * f_2(e_{dist}(v) - c_{dist}(v))
\end{equation}

Finally, we accumulate these consistency values across all frames and for all depth-consistent voxels. We compute the expected color based on the projection of the voxel and add to the corresponding HSV bin the value of the \textit{color} function, as in Equation \ref{eq:12}.
\begin{equation} \label{eq:12}
bins[v, bin(v)]_{i+1} = bins[v, bin(v)]_{i} + color(v)
\end{equation}

After passing through all the frames, normalization is applied over the bins array, such that the difference in magnitudes is taken into account across all bins.
\begin{equation} \label{eq:13}
bins[v, bin(v)] = \frac{bins[v, bin(v)]}{\sum_{bin}bins[v]}
\end{equation}

With the normalized bins array, the color consistency of a voxel is given by the maximum value across all bins, greater than a tunable threshold. We call this threshold \(\varepsilon_{hsv}\). If a voxel is color consistent, its color is therefore given by the bin corresponding to that maximum value. In order to obtain the RGB color, we compute the corresponding bins on the H, S, and V components based on the bin, pick the middle points of each interval and convert them back to the RGB color spectrum.

\subsection{Multi-scale voxel grid}\label{sec:vg}

The multi-scale paradigm implies the reconstruction of the scene with voxels varying in size and combining the end results in a way that improves the image quality when compared to the individual ones. Following this idea, we propose to use the multi-scale approach by reconstructing the same scene using multiple increasing voxel sizes and blending the resulting RGB images. Given a set of increasing sizes \(S = S_1, S_2, ..., S_k\), for each \textit{k}, the corresponding voxel grid is reconstructed, and the images are regenerated, as they are observed from the camera poses in the test set. The resulting set of images consists of multiple views of the reconstructed scene, at multiple scales, for each \textit{k}. Finally, the blending is done on an image level using a masking mechanism, starting from the smallest size \(S_1\). The mask denotes those pixels for which they are observed as empty, such that the larger scales images can fill the empty pixels in the smaller scales images. During the reconstruction process, the voxels within the scene are independent of each other, thus allowing for a better implementation of the initial voxel grid. The initial proposal described a single larger voxel grid that had a similarly large memory footprint. Due to the independent relation between the voxels, the large voxel grid can be split into disjoint smaller voxel grids, which we call \(blocks\), for which the reconstruction algorithm is applied iteratively. Moreover, as the blocks are getting smaller, the voxels comprised in a single block can also get smaller, such that the scene is reconstructed at a higher resolution.


\subsection{Voxel grid parameters} \label{params}
In this subsection, we present the parameters used for the construction of the Voxel Grid. Throughout all of our experiments, we have set the voxel size as \(voxel\_size=0.5\), which roughly translates to a size of half a meter. The grid consists of multiple voxels "stitched" together with no space in between them. Thus we arrange the voxels in a parallelepiped defined with a position in the World Space and a width, height, and length, signifying its sides. The position is, essentially, the position of one voxel in the world, be it either the central voxel of the grid or one of its corners. The width, height, length, and position are scene dependent, as the voxel grid must cover most of the scene, while also taking into account the positions and orientations of the drone. Even though this can be done in a dynamic way based on the drone information, we have opted to set them empirically and keep the sides of the voxel grid constant, while moving its position based on the reconstructed scene. The height is set to 80 meters (160 voxels), the length is set to 550 meters (1100 voxels) and the width is set to 550 meters (1100 voxels).

\subsection{Reconstruction enhancement}\label{sec:re}

Despite our efforts to maximize the density due to the image noise and low consistency score our output tends to have a number of empty regions. In order to address this problem, we developed an algorithm to densify its prediction. Similar to the depth completion algorithms, ours attempts an RGB completion based on the reconstructed input. The algorithm is trained and tested on the same sets used for voxel carving. The algorithm consists of a small U-Net~\cite{ronneberger2015u} (1M parameters) trained to fill in the gaps. Two versions of the algorithm were tested, one with RGB input and another one with RGB and pose encoded as 6 channels, 3 for normalized pose and 3 for Euler angles encoded as $sinx*cosx$.

\section{Experimental Analysis} 

We focus on aerial scene reconstruction - as opposed to a single-centered object generally targeted by fast reconstruction methods. We use the public real UFODepth dataset proposed in~\cite{licuaret2022ufo}, which features telemetry data (commonly provided by UAV manufacturers) and a diverse set of landscapes, with both vegetation and man-made structures. The flight altitude is generally 50m and the flight trajectory is manual. Although the original resolution is 4K (3180 × 2160 px), we rescale them to 1920 × 1080 for most experiments). We use 5 minutes videos (9000 frames in total) which we sample every 20 frames. 
We report results for all four scenes from the dataset - Slanic, Olanesti, Chilia, and Herculane. For the first stage of the algorithm, we extract metric depth using an optical flow-based method, termed OdoFlow from~\cite{licuaret2022ufo} and FlowFormer~\cite{huang2022flowformer} as a pretrained optical flow algorithm. For validation, we also compute the depth from SfM using Meshroom~\cite{griwodz2021alicevision}. 


We split each scene of the dataset into two parts: the reconstruction part and the reprojection part. The reconstruction part contains 80\% of the data points, leaving the rest of 20\% as the reprojection data. We reconstruct the scene based on only the reconstruction data and test the results using the reprojection data. In this way, we ensure that the results are based on data that was not used during the reconstruction process. During the reprojection phase, we use the poses of the drone to place a virtual camera with the same intrinsic parameters as the intrinsics of the drone and capture an RGB Image for each pose.

\begin{figure}
\includegraphics[scale=0.295]{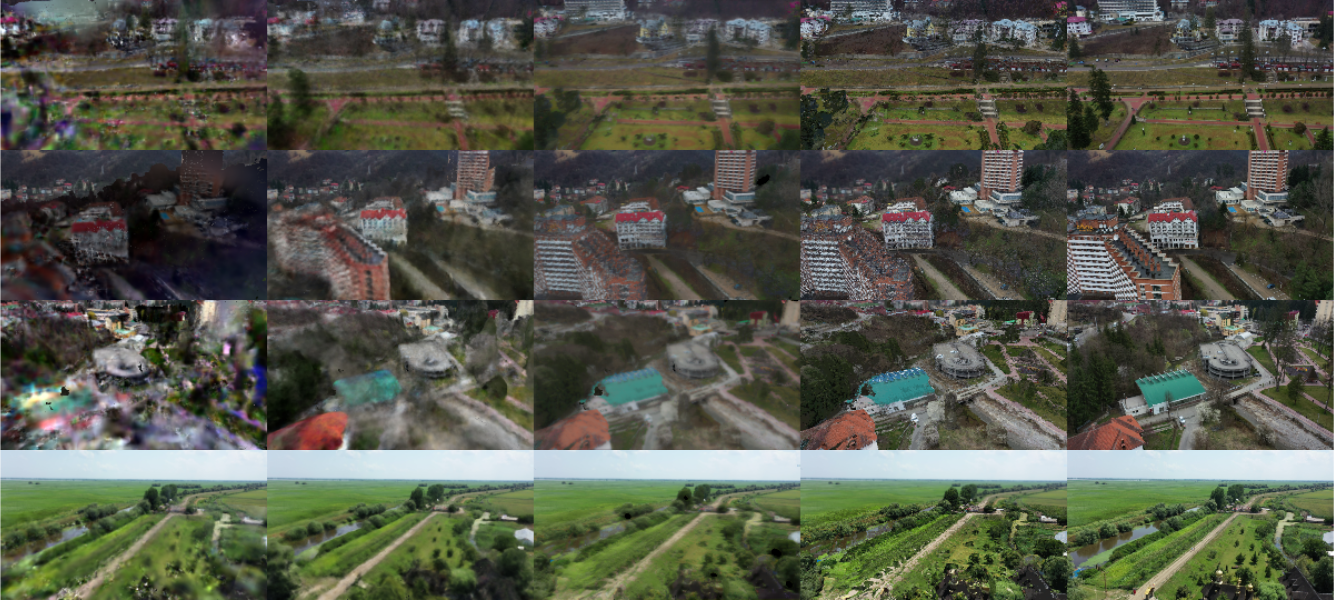}
\caption {\label{figure:qualitative_results} Qualitative testing results from all scenes. From left to right, Plenoxels\cite{yu_and_fridovichkeil2021plenoxels}, Instant-NGP\cite{muller2022instant}, MSVC full, MSVC before learning, ground truth. While other methods perform poorly on unseen images, displaying artifacts and significant noise levels, ours preserves more fine details and manages to reproduce a higher fidelity image. Best seen in color and zoom. A complementary video with more qualitative results can be seen here~\url{https://youtu.be/dqN_OUVzscE}.}
\vspace{-3mm}

\end{figure}

\begin{figure}
\includegraphics[scale=0.102]{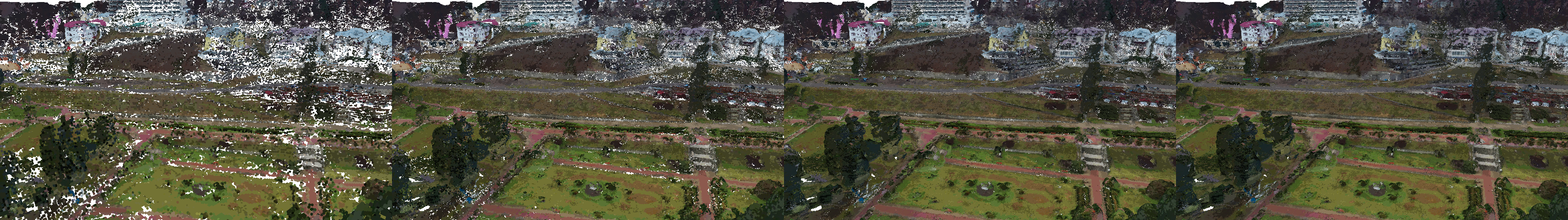}
\caption {\label{tab:multiscale_carving_and_merged} Multi-scale voxel carving reconstruction at different scales - from left to right, 0.5, 0.25, 0.125 m voxel, followed by the merged result. Multi-scaling allows us to better reconstruct the scene and reduces the void areas (white in the picture above). } \label{fig:multiscale_completion}
\end{figure}

\subsection{Results and discussion}
\label{sec:comparisons}

We compare our method against both implicit and explicit depth methods with similar computing requirements. Instant-NGP~\cite{muller2022instant} is a fast framework that benefits from a 2D location hashing scheme and a CUDA-optimized MLP architecture. We compare two versions - vanilla and depth input. The depth support was added after the initial code release and was not officially supported. We use the same flow-based depth for our method, in order to provide a fair comparison. 

We have chosen two complementary voxel-based methods. The first one, Direct Voxel Grid Optimization\cite{sun2022direct} is similar to a NeRF that has the neural network replaced by a dense voxel grid, making it the most similar to our approach. On the other hand, Plenoxels~\cite{yu_and_fridovichkeil2021plenoxels} use a sparse grid in order to build a more efficient scene representation.

\begingroup
\setlength{\tabcolsep}{3.6pt}
\begin{table*}[t]

\begin{center}
\begin{tabular}{|l|c|c|c|c|c|c|c|c|c|c|}
\hline
\backslashbox{Method}{Scene}
 & \multicolumn{2}{|c|}{ Slanic } & \multicolumn{2}{|c|}{ Olanesti } & \multicolumn{2}{|c|}{ Chilia } & \multicolumn{2}{|c|}{ Herculane }  & \multicolumn{2}{|c|}{ Mean } \\ \hline
 & Train & Test & Train & Test & Train & Test & Train & Test & Train & Test\\ \hline

DVGO\cite{sun2022direct} & 18.90 & 15.34 & 17.29 & 14.34 & 18.01 & 16.43 & 18.12 & 15.43 & 18.08 & 15.38\\ \hline

Plenoxels\cite{yu_and_fridovichkeil2021plenoxels} & \textbf{23.08} & 17.22 & \textbf{21.15} & 15.92 & 21.19 & 18.59 & 21.60 & 16.07 & 19.5 & 15.38\\ \hline

Instant-NGP\cite{muller2022instant} & 19.91 & 16.54 & 20.36 & 19.25 & \textbf{21.88} & 21.01 & 20.69 & 19.12 & \textbf{20.71} & 18.98 \\ \hline

\shortstack{Instant-NGP, Depth (1)} & 20.18 & 16.63 & 14.98 & 15.55 & 22.09 & \textbf{21.32} & \textbf{20.48} & 18.04 & 19.43 & 17.88\\ \hline \hline

MSVC, Depth (1) & 20.64 & 20.05 & 19.66 & 19.35 & 20.53  & 19.91 & 20.05 & 19.93 & 20.22  & 19.81 \\ \hline

MSVC, Depth (2) & 21.46 & \textbf{20.49} & 19.55 & \textbf{19.43} & 21.41  & 21.29 & 20.18 & \textbf{20.25} & 20.65  & \textbf{20.36} \\ \hline

\end{tabular}
\end{center}
\caption{\label{tab:results_reconstruction_scene}
PSNR reconstruction results on the UFODepth dataset. While many methods achieve better results at learning the training set, they have poorer performance for poses that are not close neighbors of the training set. Our method consistently yields better reconstruction errors on the testing set, yielding much lower performance drops on the test set. Depth (1) denotes OdoFlow depth, while (2) refers to SfM depth. Higher is better.}
\end{table*}
\endgroup

As Table \ref{tab:results_reconstruction_scene} shows, we achieve competitive performance among the compared methods on the training set and superior numbers on the testing set. It is notable that implicit methods that use input depth make little use of it, often resulting in poorer performance (e.g., Instant-NGP with depth). We argue that the testing set should be the focus for the novel view reconstruction, as overfitting the training set generally results in artifacts on novel views - missing structures or low-density structures, as shown in Figure~\ref{figure:qualitative_results}.

We conduct an ablation study on the final self-supervised learning stage and report our results in Table~\ref{tab:results_self_supervised_ablation}. This validates the final stage - it filters out the noise from the carving algorithm. Even without this later stage, our algorithm still yields competitive performance on the test scenes.


\begingroup
\setlength{\tabcolsep}{4.5pt}
\begin{table*}[t]

\begin{center}
\begin{tabular}{|c|c|c|c|c|c|c|c|c|c|c|}
\hline
\backslashbox{Method}{Scene}
 & \multicolumn{2}{|c|}{ Slanic } & \multicolumn{2}{|c|}{ Olanesti } & \multicolumn{2}{|c|}{ Chilia } & \multicolumn{2}{|c|}{ Herculane }  & \multicolumn{2}{|c|}{ Mean } \\ \hline
 & Train & Test & Train & Test & Train & Test & Train & Test & Train & Test\\ \hline

\shortstack{MSVC Steps 1-2} & 19.01 & 18.33 & 17.18 & 17.06 & 18.80  & 18.72 & 18.68 & 18.48 & 18.41  & 18.14 \\ \hline

\shortstack{MSVC full} & 21.46 & \textbf{20.49} & 19.55 & \textbf{19.43} & 21.41  & 21.29 & 20.18 & \textbf{20.25} & 20.65  & \textbf{20.36} \\ \hline

\end{tabular}
\end{center}
\caption{\label{tab:results_self_supervised_ablation}
Ablation study for the self-supervised learning module. The result from the multi-scale voxel grid is used as a prior for the student, using only the training data available for the voxel carving algorithm (steps 1-2). This final step enables higher quality novel view synthesis on the testing set (full pipeline, steps 1-3). Higher is better.}
\end{table*}
\endgroup

\begingroup
\setlength{\tabcolsep}{4.5pt}
\begin{table}[h!]
\begin{center}
\begin{tabular}{|c|c|c|c|}
\hline
Voxel Size[m] & Projection[s] & Colorization[s]& Total[s] \\ \hline
0.125 & 2.07 & 4.55 & 6.62\\ \hline
0.25 & 0.35 & 1.11 & 1.46\\ \hline
0.5 & 0.07 & 0.12 & 0.19\\ \hline
\end{tabular}
\end{center}
\caption{\label{tab:results_time}
Mean running times of the reconstruction algorithm per training image. Colorization tends to take most of the time when decreasing the voxel size due to the color consistency algorithm. Although a finer voxel size than 0.125m is possible, it was deemed too compute-intensive and delivers small performance gains.}
\vspace{-3mm}
\end{table}
\endgroup

\noindent\textbf{Runtime}. In terms of technical performance, we have measured the running time of the algorithm on a single Tesla P100 GPU, on a machine equipped with an Intel Xeon E5-2640 v4 CPU. The projection was run on the GPU to leverage fast matrix multiplication, while the colorization is executed on the CPU due to the large memory footprint of the bins array. We describe the mean running times in Table~\ref{tab:results_time}, where the reported times are averaged across all scenes with voxels varying in size. The scene consists of a grid of voxel blocks, where each block has a height of 80m, width of 50m, and length of 50m. As each block is reconstructed independently, the running time increases linearly with the number of blocks used in the initial grid.


\noindent\textbf{Impact.} Novel view synthesis with explicit depth representation is a helpful tool for safer navigation or environmental monitoring. With the ever-increasing resolution capabilities of UAV cameras, our multi-resolution grid could be adapted to smaller voxel sizes and result in more accurate maps. 

\section{Conclusion}

We present a method for novel view synthesis of real scenes which is grounded in the real world and proved its effectiveness on large-scale scenes with noisy pose measurements. While most methods are tested on synthetic scenes and on the same images used for training, we evaluated and showed significant improvements over state-of-the-art from viewpoints that are drastically different from the ones seen in training. Moreover, we showed that our method suffers from very little degradation between the seen (training) vs unseen (testing), unlike the recent competition which shows significant degradation between the two, indicating strong overfitting. On the algorithmic side, we demonstrated the effectiveness of both modules. Mlti-scale voxelization ensures that we start from good coverage and low error from drastically different unseen views, while our reconstruction enhancement module is able to self-train on the seen views and effectively adapt to each seen in a relatively short amount of time. We believe that our dual geometric and deep learning does cover new ground in this area and has the potential to push the boundaries further towards novel view synthesis that is grounded in reality.

\section{Acknowledgements}

This work was funded by UEFISCDI, under Projects EEA-RO-2018-0496 and PN-III-P4-ID-PCE-2020-2819. We want to express our sincere gratitude towards Aurelian Marcu and The Center for Advanced Laser Technologies (CETAL) for their generosity and for providing us access to GPU computational resources.

{
\bibliographystyle{ieee_fullname}
\bibliography{egbib}
}

\end{document}